\documentclass[10pt,twocolumn,letterpaper]{article}
\usepackage{CJKutf8}
\usepackage[ruled,vlined,linesnumbered]{algorithm2e} 
\usepackage{xcolor}                    
\usepackage{amsmath,amssymb}
\usepackage{amsthm}
\usepackage{bm}         
\usepackage{graphicx}   
\usepackage{geometry}   
\usepackage{marvosym} 
\usepackage{adjustbox}
\usepackage{booktabs}   
\usepackage{multirow}   
\usepackage{colortbl}   
\usepackage{array}      
\usepackage{booktabs}   
\usepackage{tabularx}   
\usepackage{amssymb}
\allowdisplaybreaks[2] 
\newcommand{\Hook}[1]{\hfill\mbox{\normalfont$\triangleright$~#1}}
\newcommand{\EndCluster}{\textnormal{end}\Hook{Cluster Stage}}
\newcommand{\EndLeaf}{\textnormal{end}\Hook{Leaf Stage}}
\SetKwFor{ForC}{for \textbf{(\textit{in parallel})}}{\ \normalfont do}{\EndCluster}
\SetKwFor{ForEachC}{for \textbf{(\textit{in parallel})}}{\ \normalfont do}{\EndLeaf}

\newtheorem{theorem}{Theorem}

\newtheorem{corollary}{Corollary}

\newtheorem{assumption}{Assumption}

\newcommand{\Fcal}{\mathcal{F}}
\newcommand{\Hcal}{\mathcal{H}}
\newcommand{\Gcal}{\mathcal{G}}
\newcommand{\disc}{\operatorname{disc}} 
\newcommand{\Rehat}{\widehat{\mathfrak{R}}} 

\usepackage[most]{tcolorbox}

\definecolor{CoolGreen}{RGB}{50,180,140}

\tcbset{
  graytheorem/.style={
    enhanced,                        
    breakable,                       
    colback=gray!12,                 
    colframe=black!80,               
    boxrule=0.6pt,                   
    arc=2pt,                         
    left=6pt, right=6pt, top=6pt, bottom=6pt,
    before skip=10pt, after skip=10pt,
    toprule at break=1pt,            
    bottomrule at break=1pt          
  }
}

\usepackage{cvpr}              

\definecolor{cvprblue}{rgb}{0.21,0.49,0.74}
\usepackage[pagebackref,breaklinks,colorlinks,allcolors=cvprblue]{hyperref}

\def\paperID{} 
\def\confName{CVPR}
\def\confYear{2026}
\title{HiLoRA: Hierarchical Low-Rank Adaptation for Personalized Federated Learning}
\author{
Zihao Peng$^{1}$, Nan Zou$^{1}$, Jiandian Zeng$^{1}$, Guo Li$^{1}$, Ke Chen$^{1}$, Boyuan Li$^{2}$, and Tian Wang$^{1, \text{\Letter}}$ \\
$^{1}$Beijing Normal University, $^{2}$Zhengzhou University \\
{\ttfamily\small \{pzh\_cs, 202311109042, liguo, kechen\}@mail.bnu.edu.cn} \\
{\ttfamily\small l202311841010602@gs.zzu.edu.cn,} 
{\ttfamily\small \{jiandian, tianwang\}@bnu.edu.cn}
}

\begin{document}
\begin{CJK}{UTF8}{gbsn}     
\maketitle
{
  \renewcommand{\thefootnote}{} 
\footnotetext{\textsuperscript{\Letter}Corresponding Author.}
}
\addtocounter{footnote}{-1} 
\begin{abstract}
Vision Transformers (ViTs) have been widely adopted in vision tasks due to their strong transferability. In Federated Learning (FL), where full fine-tuning is communication-heavy, Low-Rank Adaptation (LoRA) provides an efficient and communication-friendly way to adapt ViTs. However, existing LoRA-based federated tuning methods overlook latent client structures in real-world settings, limiting shared representation learning and hindering effective adaptation to unseen clients. To address this, we propose HiLoRA, a hierarchical LoRA framework that places adapters at three levels: root, cluster, and leaf, each designed to capture global, subgroup, and client-specific knowledge, respectively. Through cross-tier orthogonality and cascaded optimization, HiLoRA separates update subspaces and aligns each tier with its residual personalized objective. In particular, we develop a LoRA-Subspace Adaptive Clustering mechanism that infers latent client groups via subspace similarity analysis, thereby facilitating knowledge sharing across structurally aligned clients. Theoretically, we establish a tier-wise generalization analysis that supports HiLoRA’s design. Experiments on ViT backbones with CIFAR-100 and DomainNet demonstrate consistent improvements in both personalization and generalization.
\end{abstract}

\vspace{-3mm}
\section{Introduction}
\label{sec:intro}

Federated Learning (FL) enables collaborative training across distributed clients without sharing raw data \cite{mcmahan2017communication}. With the rapid progress of large-scale Foundation Models (FMs), combining FL and FMs has become a compelling paradigm \cite{chen2024feddat,zhuang2023foundation,10944288}. Pretrained FMs provide powerful initialization that accelerates convergence and enhances generalization, while FL alleviates the data-sharing bottleneck of cross-domain data collection. 

\begin{figure}
\centering
\includegraphics[scale=0.72]{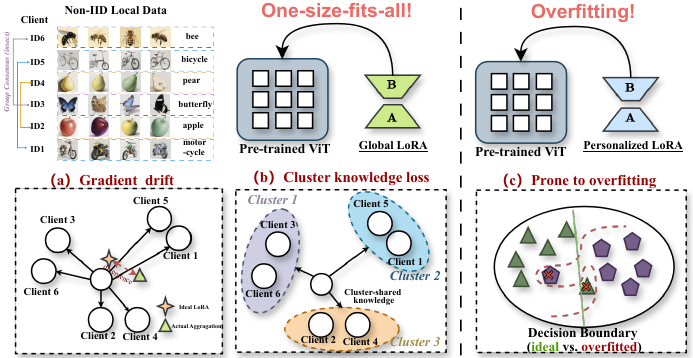}
\caption{Challenges of \textbf{Dual-LoRA}.
Six clients own non-IID data that form three latent clusters (vehicles, insects, fruits).
Global LoRA forces a “one-size-fits-all” adapter, causing (a) gradient drift and (b) loss of cluster-level cues.
Fully personalized LoRA removes sharing but (c) overfits scarce local data.
These limitations motivate a hierarchical LoRA design.}
\vspace{-5.0mm}
\label{fig1}

\end{figure}

Building upon these advantages, recent research has increasingly explored how this collaboration between FL and FMs can be leveraged in specific domains. Among them, Computer Vision (CV) has emerged as one of the most promising areas, owing to the vast amount of visual data distributed across diverse devices and institutions. In CV, Vision Transformers (ViTs) are representative foundation models that transfer across diverse downstream tasks
\cite{kirillov2023segment,zhu2021deformable,han2022survey}.  However, full fine-tuning of ViTs in FL incurs prohibitive communication overhead \cite{10930890,10944288,liu2026fedadamwcommunicationefficientoptimizerconvergence}, motivating Parameter-Efficient Fine-Tuning (PEFT). Among PEFT methods, Low-Rank Adaptation (LoRA) \cite{hu2022lora,zhang2023adaptive} achieves performance close to full fine-tuning while  reducing communication costs, making it a promising solution for efficient FM adaptation in FL \cite{10666083,bai2024federated,peng2025fedhl,wang2024flora}. Although PEFT enables large FMs to be deployed in FL, the long-standing tension between global utility and local personalization remains unresolved \cite{meng2024improving,tan2022towards,guo2023out}. Under the prevalent non-IID setting, a single global model neither generalizes well to diverse client distributions nor adapts to client-specific patterns, leading to client drift and degraded performance \cite{sun2023fedperfix,guo2023pfedprompt}. This gap requires an adaptation mechanism that can maintain PEFT-level efficiency while reconciling personalization and generalization in FM-driven FL systems.

Recent studies extend PEFT to FL by equipping each client with two distinct LoRA adapters, one aggregated globally and the other tuned locally \cite{yi2023pfedlora,long2024dual}, to account for both personalization and generalization, as illustrated in Fig.~\ref{fig1}. Although this “dual-adapter” paradigm improves upon fully global or fully local LoRA, it still has some inherent drawbacks. First, severe data heterogeneity causes clients to pursue divergent local optima, producing gradient drift that pulls the aggregated global adapter away from the optimal global solution \cite{liuimproving,gao2022feddc}.  Second, local adapters trained only on limited and skewed client data tend to overfit, leading to brittle decision boundaries and poor in-client generalization. Third, dual-adapter designs neglect the latent group structures inherent in real-world FL deployments, where different clients may naturally form subgroups due to data characteristics. Thus, the dispersal of subgroup-shared knowledge is either diluted into the global adapter or trapped within local ones, impeding transfer across related clients \cite{ma2023structured,ghosh2020efficient}.  Consequently, group-specific information remains under-exploited, and the trade-off between generalization and personalization is still coarse-grained, leaving potential performance gains unrealized.

To address these limitations, we introduce HiLoRA, a hierarchical low-rank adaptation framework for federated foundation models. HiLoRA organizes LoRA adapters into three hierarchical tiers: root, cluster, and leaf levels, and enforces cross-tier orthogonality to disentangle global and personalized knowledge. (1) The Root-LoRA serves as the globally shared adapter, capturing common patterns across all clients and providing the foundation for subsequent cluster and leaf adaptations. (2) The Cluster-LoRA modules are shared among clients with similar data distributions. To identify such client communities, we propose a \emph{LoRA-Subspace Adaptive Clustering} method that automatically partitions clients into clusters, each associated with a shared adapter.  By explicitly modeling these latent client clusters, the cluster tier captures subgroup-shared representations and aggregates within clusters, thereby mitigating heterogeneity-induced gradient drift and promoting consistent learning dynamics. (3) The Leaf-LoRA is unique to each client, capturing residual patterns specific to each client’s data, with regularized updates on local data that alleviate overfitting and improve in-client generalization.  Finally, we conduct a generalization analysis to demonstrate that clustering and cross-tier orthogonality jointly tighten the upper bound on client-level excess risk, thereby providing theoretical support for HiLoRA’s hierarchical design. Our main contributions are summarized as follows:
\begin{itemize}
    \item We introduce HiLoRA, a novel hierarchical low-rank adaptation framework for FL that organizes LoRA adapters into root, cluster, and leaf tiers, with each tier having its own personalized objective.
    
    \item We develop a clustering mechanism based on LoRA subspace similarity, which infers latent client groups via principal-angle analysis and enables knowledge sharing among aligned clients within each cluster.

    \item We provide a generalization analysis for HiLoRA, showing that orthogonality mitigates cross-tier interference and that clustering reduces distributional discrepancies, jointly yielding tighter generalization bounds.
    
    \item Extensive experiments on the CIFAR-100 and DomainNet datasets show that HiLoRA outperforms state-of-the-art methods in both personalization and generalization.
\end{itemize}

\section{Preliminaries}
\label{ssec:pre_and_pro}
\subsection{LoRA Recap}  
Following \cite{hu2022lora,dettmers2023qlora}, a pre-trained weight matrix $\mathbf{W}_{0}\in\mathbb{R}^{p\times q}$ is frozen, and the adapted weight $\mathbf W$ is obtained by adding a learnable low-rank update $\Delta\mathbf W$ parameterized as the product of two factors:
\begin{equation}
\label{eq:lora}
\mathbf W = \mathbf W_0 + \Delta\mathbf W, 
\quad \Delta\mathbf W = \mathbf B \mathbf A.
\end{equation}
Here, $\mathbf{B}$ and $\mathbf{A}$ are learned low-rank factors with $\mathbf{B}\in\mathbb{R}^{p\times r}$ and $\mathbf{A}\in\mathbb{R}^{r\times q}$, where $r\ll \min\{p,q\}$. Since $r$ is much smaller than the dimensions of $\mathbf{W}_{0}$, the number of trainable parameters is substantially reduced. In addition, the additive form in \cref{eq:lora} naturally lends itself to extensions involving multiple LoRA modules, leading to a hierarchical composition that can be formally expressed as:
\begin{equation}
\label{eq:hlora}
\mathbf{W}=\mathbf{W}_0+\sum_{h=1}^{H}\Delta\mathbf{W}_{h},
\end{equation}
with $h\in \{1,\dots,H\}$ indexing hierarchy levels and $\Delta\mathbf{W}_h$ denoting the update at level $h$.

\subsection{Three-Level LoRA}  
Building on the hierarchical formulation in \cref{eq:hlora}, 
we introduce a three-level structure over the $N$ clients, consisting of root, cluster, and leaf levels. The clients are partitioned into $K$ disjoint clusters $\{\mathcal C_j\}_{j=1}^K$, 
and adapters are attached at each level as follows:  
\begin{itemize}[leftmargin=*,nosep]
  \item \emph{Root:} a single adapter $(\mathbf{B}_{r},\mathbf{A}_{r})$ shared by all $N$ clients;
 \item \emph{Cluster:} adapters $\{(\mathbf{B}_{c,j},\mathbf{A}_{c,j})\}_{j=1}^{K}$, each derived from the root and shared among clients in the cluster $\mathcal{C}_j$;
\item \emph{Leaf:} adapters $\{(\mathbf{B}_{\ell,i},\mathbf{A}_{\ell,i})\}_{i=1}^{N}$, each associated with the cluster $\mathcal{C}_{j(i)}$ of client $i$, and unique to client $i$.
\end{itemize}
For client $i$ in cluster $\mathcal{C}_{j(i)}$, the effective LoRA update is
\begin{equation}
\label{eq:three_levels}
\Delta\mathbf{W}_{i}
=\mathbf{B}_{r}\mathbf{A}_{r}
+\mathbf{B}_{c,j(i)}\mathbf{A}_{c,j(i)}
+\mathbf{B}_{\ell,i}\mathbf{A}_{\ell,i},
\end{equation}
where $j(i)\in\{1,\dots,K\}$ denotes the cluster assignment of client $i$. 
The three terms correspond respectively to $(\mathbf{B}_{r},\mathbf{A}_{r})$ as a global backbone, 
$(\mathbf{B}_{c,j(i)},\mathbf{A}_{c,j(i)})$ capturing intra-cluster commonality, 
and $(\mathbf{B}_{\ell,i},\mathbf{A}_{\ell,i})$ modeling client-specific personalization.

\subsection{Federated Learning Setup}
We consider a personalized Federated Learning (pFL) setup with $N$ clients.
Client $i$ holds a private dataset $\mathcal{D}_i=\{(\mathbf{x}_i^{(k)}, y_i^{(k)})\}_{k=1}^{n_i}$ that remains on-device, where $n_i=|\mathcal{D}_i|$.
Each input $\mathbf{x}\in\mathbb{R}^d$ has a label $y\in\{1,\dots,C\}$, and the predictor is $f(\cdot;\mathbf{W}):\mathbb{R}^d\to\mathbb{R}^C$.
For client $i$, the personalized model is written as $\mathbf{W}_i=\mathbf{W}_0+\Delta\mathbf{W}_i$.
The local empirical risk is
$
\ell_i(\mathbf{W}_i)=\frac{1}{n_i}\sum_{k=1}^{n_i} L\!\big(f(\mathbf{x}_i^{(k)};\mathbf{W}_i),\, y_i^{(k)}\big)
$, where $L$ denotes the cross-entropy loss.

Unlike standard FL, which optimizes a single shared model $\mathbf{W}$ via $\min_{\mathbf{W}}\sum_{i=1}^N \pi_i\,\ell_i(\mathbf{W})$, the pFL formulation optimizes client-specific models $\{\mathbf{W}_i\}_{i=1}^N$ that share a hierarchical LoRA parameterization. Specifically, for client $i$, we denote by
$
\mathcal{A}_i(\Phi)=\big\{(\mathbf B_r,\mathbf A_r),\ (\mathbf B_{c,j(i)},\mathbf A_{c,j(i)}),\ (\mathbf B_{\ell,i},\mathbf A_{\ell,i})\big\}
$
the subset on its root–cluster–leaf path, and compose the effective model as $\mathbf W_i(\Phi)=\mathbf W_0+\sum_{(\mathbf B,\mathbf A)\in \mathcal{A}_i(\Phi)} \mathbf B\,\mathbf A$. We consider the following aggregated objective:
\begin{equation}
\label{eq:pfed_obj}
\min_{\Phi}\ \sum_{i=1}^N \pi_i\,\ell_i\!\big(\mathbf{W}_i(\Phi)\big),
\end{equation}
where $\pi_i=\tfrac{n_i}{\sum_{u=1}^N n_u}$ are data-proportional weights. Each client performs local updates and periodically communicates with the server without sharing raw data.

\section{The HiLoRA Framework}
There are three fundamental challenges in FL with hierarchical low-rank adaptation:

\begin{itemize}

  \item \textbf{Knowledge hierarchy (Q1):} How can we disentangle global consensus, cluster-level commonalities, and client-specific patterns within a unified low-rank representation?
  
 \item \textbf{Adaptive clustering (Q2):} How can we discover latent client structures from LoRA updates without pre-defined groups and promote effective mutual knowledge sharing?

  \item \textbf{Tier-wise training (Q3):} How can we design a cascaded, tier-wise optimization scheme for training root–cluster–leaf LoRA adapters that achieves hierarchical personalization?
\end{itemize}

\begin{figure*}[!t]
\centering
\includegraphics[width=\linewidth]{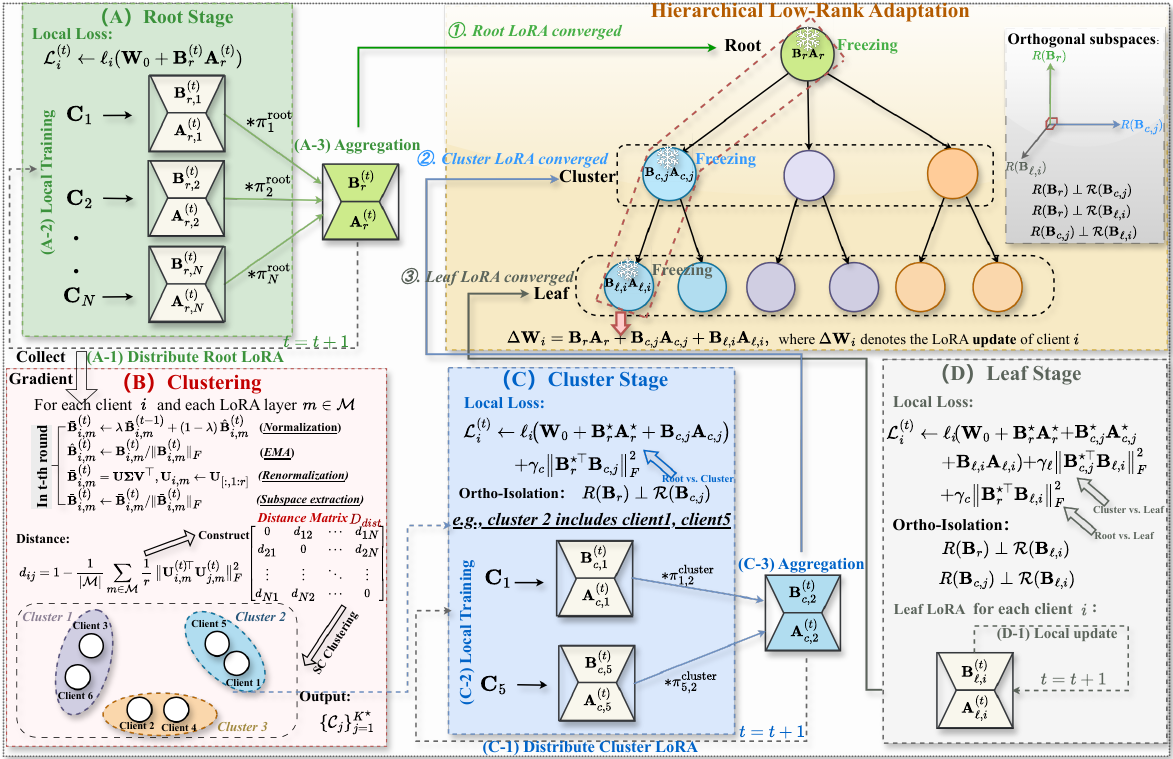}
\caption{\textbf{HiLoRA overview.} \emph{Top-right:} cascaded tier-wise optimization with progressive freezing. (A) Train the \textcolor{ForestGreen}{Root-LoRA} as the global adapter; (B) Perform LoRA-Subspace Adaptive Clustering to identify client communities. (C) Update the \textcolor{RoyalBlue}{Cluster-LoRA} with an orthogonality constraint to the frozen root; (D) Adapt the \textcolor{Gray}{Leaf-LoRA} orthogonal to both root and cluster tiers. Each client $i$ with cluster index $j=j(i)$ updates its effective LoRA as 
$\Delta\mathbf{W}_{i}=\mathbf{B}_{r}\mathbf{A}_{r}+\mathbf{B}_{c,j}\mathbf{A}_{c,j}+\mathbf{B}_{\ell,i}\mathbf{A}_{\ell,i}$.}
\label{fig2}
\vspace{-2mm}
\end{figure*}

HiLoRA tackles these challenges by combining orthogonal low-rank decomposition (\textbf{Q1}), adaptive clustering in the LoRA-update subspace (\textbf{Q2}), and cascaded tier-wise optimization with cross-tier orthogonality (\textbf{Q3}) in a unified workflow, as illustrated in Fig.~\ref{fig2}.

\subsection{Hierarchical Orthogonal LoRA Decomposition}
HiLoRA models knowledge at multiple granularities by decomposing each client's weight update \(\Delta\mathbf{W}_i \in \mathbb{R}^{p \times q}\) into three orthogonally constrained low-rank components:
\begin{equation}
  \Delta\mathbf{W}_i
  = \mathbf{B}_r\mathbf{A}_r
    + \mathbf{B}_{c,j(i)}\mathbf{A}_{c,j(i)}
    + \mathbf{B}_{\ell,i}\mathbf{A}_{\ell,i}.
\end{equation}
Here, each basis matrix $\mathbf{B} \in \mathbb{R}^{p \times r}$ specifies the direction of adaptation, i.e., \emph{where to adapt}, while its paired coefficient matrix $\mathbf{A} \in \mathbb{R}^{r \times q}$ controls the magnitude, i.e., \emph{how much to adapt}, along those directions.
To prevent different tiers from adapting along overlapping directions, we aim to enforce pairwise orthogonality on their column spaces for each path \(r\!\to\! j(i)\!\to\! i\):
\begin{equation}\ 
\forall\, U\neq V \in \{\mathcal R(\mathbf B_r),\mathcal R(\mathbf B_{c,j(i)}),\mathcal R(\mathbf B_{\ell,i})\}:\ U\perp V,
\end{equation}
where \(\mathcal R(\mathbf B)\subseteq\mathbb R^p\) denotes the column space of \(\mathbf B\). 

By encouraging orthogonality among the basis matrices $\mathbf B$, we ensure that each tier tends to isolate its intended residual component. In particular, $\mathbf B_r$ provides \textit{global} adaptation directions and $\mathbf A_r$ scales them to encode trends shared across all clients, while $\mathbf B_{c,j}$ defines shared \textit{cluster-level} directions and $\mathbf A_{c,j}$ weights them to capture intra-cluster commonalities. Finally, $\mathbf B_{\ell,i}$ supplies \textit{client-specific} directions and $\mathbf A_{\ell,i}$ modulates them to represent residuals not explained by higher tiers. This decoupling of direction and magnitude allows HiLoRA to naturally support global, cluster-level, and client-specific personalization.

\subsection{LoRA-Subspace Adaptive Clustering}
\label{sec:gac}
To identify client groups without exposing raw data, we cluster clients based on the similarity of their LoRA subspaces. Specifically, for each client $i$ in communication round $t$, we extract the LoRA basis $\mathbf B^{(t)}_{i}$, which encodes its adaptation directions. The basis is subsequently normalized to a unit Frobenius norm $\hat{\mathbf B}^{(t)}_{i}=\mathbf B^{(t)}_{i}/\|\mathbf B^{(t)}_{i}\|_F$, and stabilized across rounds via an EMA with decay $\lambda\in(0,1)$:
\begin{equation}
\bar{\mathbf{B}}^{(t)}_{i}=\lambda\,\bar{\mathbf{B}}^{(t-1)}_{i}+(1-\lambda)\,\hat{\mathbf{B}}^{(t)}_{i},\quad
\bar{\mathbf{B}}^{(t)}_{i}\leftarrow \frac{\bar{\mathbf{B}}^{(t)}_{i}}{\|\bar{\mathbf{B}}^{(t)}_{i}\|_F}.
\end{equation}
We then compute the SVD of $\bar{\mathbf{B}}^{(t)}_{i}$ and retain the top-$r$ left singular vectors $\mathbf{U}^{(t)}_{i}\in \mathbb{R}^{p \times r}$, which span the subspace of adaptation directions and provide reparameterization invariance and noise robustness. For a single module, the pairwise distance is defined as follows:
\begin{equation}
d_{ij}
= 1 - \frac{1}{r}\sum_{s=1}^{r}\cos^{2}\theta_s
= 1 - \frac{1}{r}\,\big\|\mathbf{U}_i^{(t)\!\top}\mathbf{U}_j^{(t)}\big\|_{F}^{2},
\end{equation}
where $\theta_s$ is the $s$-th principal angle whose cosine equals the $s$-th singular value of $\mathbf{U}_i^{(t)\!\top}\mathbf{U}_j^{(t)}$. If multiple LoRA-instrumented layers are used, $d_{ij}$ is computed for each layer and then averaged over the layers. Collecting all $d_{ij}$ yields the client–client distance matrix:\begin{equation}
\mathbf D_{\text{dist}} \;=\;
\begin{bmatrix}
0        & d_{12} & \cdots & d_{1N} \\
d_{21}   & 0      & \cdots & d_{2N} \\
\vdots   & \vdots & \ddots & \vdots \\
d_{N1}   & d_{N2} & \cdots & 0
\end{bmatrix}.
\end{equation} Finally, the matrix \(\mathbf{D}_{\text{dist}}\) is converted into an affinity matrix \(\mathbf{S}\) using a Gaussian kernel \(S_{ij}=\exp\!\big(-d_{ij}^2/(2\sigma^2)\big)\) with \(S_{ii}=1\) and \(\sigma\) set to the median off-diagonal distance, and then proceed to run Spectral Clustering (SC) on \(\mathbf{S}\) \cite{ng2001spectral}. The number of clusters is chosen by sweeping $K\in[K_{\min},K_{\max}]$ and selecting the maximizer of the eigengap of the normalized Laplacian spectrum \cite{von2007tutorial}, yielding $\{\mathcal{C}_j\}_{j=1}^{K^\star}$, which is then used in the subsequent cluster stage of the cascaded optimization.

Intuitively, aligning clients that have similar adaptation subspaces $\mathcal R(\mathbf B)$ strengthens within-cluster knowledge sharing and suppresses cross-cluster interference during later optimization stages. In contrast, non-hierarchical methods treat clients in isolation, hindering targeted sharing and amplifying cross-client negative transfer.

\subsection{Cascaded Tier-wise Optimization}
We train the adapters in a cascade: first the root, then the cluster, and finally the leaf. At each stage, we optimize the active tier on its designated data while freezing the previously learned adapters, so subsequent stages emphasize directions that are complementary to the frozen ones. 
In each stage, we minimize the objective for \cref{eq:pfed_obj} over the active tier’s adapters while keeping the others fixed, and we terminate that stage once a standard stopping criterion is met~\cite{nocedal2006numerical}. 
To reduce cross-tier redundancy, we apply orthogonality regularization to the basis matrices $\{\mathbf B_r,\mathbf B_{c,j(i)},\mathbf B_{\ell,i}\}$ along each path of each client $i$ from root to cluster to leaf, which discourages reuse of directions and mitigates interference.

\noindent\textbf{Root stage.}
We learn a single global root adapter $(\mathbf B_r,\mathbf A_r)$ shared by all clients by minimizing the weighted loss over all $N$ clients:
\begin{equation}
(\textcolor{ForestGreen}{\mathbf B_r^\star},\ \textcolor{ForestGreen}{\mathbf A_r^\star})
= \arg\min_{\textcolor{ForestGreen}{\mathbf B_r},\ \textcolor{ForestGreen}{\mathbf A_r}}
\ \sum_{i=1}^N \pi^{\text{root}}_i\,
\ell_i\!\big(\mathbf W_0+\textcolor{ForestGreen}{\mathbf B_r}\,\textcolor{ForestGreen}{\mathbf A_r}\big),
\end{equation}
where $\pi_i^{\text{root}}=\tfrac{n_i}{\sum_{u=1}^N n_u}$ denotes the aggregation weight for client $i$. At round $t$, each client $i$ uses $(\mathbf B_r^{(t)},\mathbf A_r^{(t)})$ as initialization and
performs a local update on its dataset $\mathcal D_i$ to obtain $(\mathbf B_{r,i}^{(t)},\mathbf A_{r,i}^{(t)})$. The server aggregates in product space:
$\;
\Delta \mathbf{W}_{r}^{(t+1)}
=\sum_{i=1}^{N}\pi_i^{\text{root}}\ \mathbf B_{r,i}^{(t)}\mathbf A_{r,i}^{(t)},$
which avoids the unwanted cross terms caused by averaging $\mathbf B$ and $\mathbf A$ separately~\cite{bai2024federated,singhal2025fedex,peng2025fedhl}. It then computes the truncated SVD to rank $r$,
$\Delta\mathbf W_r^{(t+1)}=\mathbf{U_r}\mathbf{\Sigma_r} \mathbf{V_r}^\top$,
and sets $\mathbf B_r^{(t+1)}:=\mathbf{U_r},\ \mathbf A_r^{(t+1)}:=\mathbf{\Sigma_r} \mathbf{V_r}^\top$. Training proceeds for at most $T_{\text{root}}$ rounds, or until the relative step-size stopping criterion is met~\cite{nocedal2006numerical}, e.g.,
$
\rho_t=\frac{\big\|\Delta \mathbf W_r^{(t+1)}-\Delta \mathbf W_r^{(t)}\big\|_F}{\big\|\Delta \mathbf W_r^{(t)}\big\|_F+\varepsilon}\ \le\ \tau_{\mathrm{rel}}
$. If triggered, we freeze $\mathbf B_r^\star:=\mathbf B_r^{(t+1)}$, $\mathbf A_r^\star:=\mathbf A_r^{(t+1)}$. 

\noindent\textbf{Cluster stage.} Given the clusters $\{\mathcal C_j\}_{j=1}^{K^\star}$ produced by the LoRA-Subspace Adaptive Clustering in Sec.~\ref{sec:gac} and the frozen root $(\mathbf B_r^\star,\mathbf A_r^\star)$, cluster $j$ solves:
\begin{equation}
\begin{aligned}
(\textcolor{RoyalBlue}{\mathbf B_{c,j}^\star},\textcolor{RoyalBlue}{\mathbf A_{c,j}^\star}) 
&= \arg\min_{\textcolor{RoyalBlue}{\mathbf B_{c,j}},\textcolor{RoyalBlue}{\mathbf A_{c,j}}}\ 
\sum_{i\in\mathcal C_j} \pi^{\text{cluster}}_{i,j}\,\Big[ 
\ell_i\!\Big(\mathbf W_0+\textcolor{ForestGreen}{\mathbf B_r^{\star}\mathbf A_r^{\star}} 
\\ 
&\quad\,+\!\textcolor{RoyalBlue}{\mathbf B_{c,j}}\,\textcolor{RoyalBlue}{\mathbf A_{c,j}}\Big) 
\,+\, \gamma_c\,\big\|\textcolor{ForestGreen}{\mathbf B_r^{\star\top}}\,\textcolor{RoyalBlue}{\mathbf B_{c,j}}\big\|_F^{2}\Big].
\end{aligned} 
\end{equation}
Here $\pi^{\text{cluster}}_{i,j}=\frac{n_i}{\sum_{u\in\mathcal C_j} n_u}$ is the aggregation weight, and $\gamma_c>0$ controls the regularization strength, encouraging $\mathcal R(\mathbf B_{c,j})\!\perp\!\mathcal R(\mathbf B_r^\star)$ to reduce cross-tier interference. At round $t$, cluster $j$ aggregates the client-level factors within $\mathcal C_j$ in product space:
$
\Delta \mathbf W^{(t)}_{c,j}=\sum_{i\in\mathcal C_j}\pi^{\text{cluster}}_{i,j}\big(\mathbf B^{(t)}_{c,i}\mathbf A^{(t)}_{c,i}\big),
$
and apply the same relative step-size criterion as in the root stage to obtain $(\mathbf B_{c,j}^{\star},\mathbf A_{c,j}^{\star})$, after which cluster $j$ is frozen.

\noindent\textbf{Leaf stage.}
With $(\mathbf B_r^{\star},\mathbf A_r^{\star})$ and $(\mathbf B_{c,j}^{\star},\mathbf A_{c,j}^{\star})$ frozen, client $i\in\mathcal C_j$ learns $(\mathbf B_{\ell,i},\mathbf A_{\ell,i})$ on $\mathcal D_i$ with cross-tier penalties:
\begin{equation}
\begin{aligned}
&(\textcolor{Gray}{\mathbf B_{\ell,i}^\star},\,\textcolor{Gray}{\mathbf A_{\ell,i}^\star}) 
= \arg\min_{\textcolor{Gray}{\mathbf B_{\ell,i}},\,\textcolor{Gray}{\mathbf A_{\ell,i}}}\; 
\ell_i\!\Big(\mathbf W_0
+\textcolor{ForestGreen}{\mathbf B_r^{\star}\mathbf A_r^{\star}} 
+\textcolor{RoyalBlue}{\mathbf B_{c,j}^{\star}\mathbf A_{c,j}^{\star}} 
\\ 
&\quad\quad+\textcolor{Gray}{\mathbf B_{\ell,i}}\,\textcolor{Gray}{\mathbf A_{\ell,i}}\Big) 
+\ \gamma_c\big\|\textcolor{ForestGreen}{\mathbf B_r^{\star\top}}\,\textcolor{Gray}{\mathbf B_{\ell,i}}\big\|_F^{2} 
+\ \gamma_\ell\big\|\textcolor{RoyalBlue}{\mathbf B_{c,j}^{\star\top}}\,\textcolor{Gray}{\mathbf B_{\ell,i}}\big\|_F^{2}.
\end{aligned}
\end{equation}
Here $j=j(i)$ is the cluster index of client $i$, and $\gamma_c,\gamma_\ell>0$ control the cross-tier orthogonality so that the leaf captures only client-specific residuals. Training halts under the same relative step-size criterion, and $({\mathbf B_{\ell,i}^\star},{\mathbf A_{\ell,i}^\star})$ is frozen.

\subsection{Personalization and Generalization in HiLoRA} 
We now discuss how this design facilitates personalization for participating clients and generalization to new clients.

\noindent\textbf{Personalized path.}  
Personalization is achieved by composing only the adapters along client $i$’s route 
\(\mathcal{A}_i(\Phi)=\{(\mathbf B_r^\star,\mathbf A_r^\star),(\mathbf B^\star_{c,\,j(i)},\mathbf A^\star_{c,\,j(i)}),(\mathbf B^\star_{\ell,\,i},\mathbf A^\star_{\ell,\,i})\}\),
yielding the effective model
\(\mathbf W_i=\mathbf W_0+\sum_{(\mathbf B,\mathbf A)\in\mathcal A_i(\Phi)}\mathbf B\,\mathbf A\).
This path-specific assembly ensures cluster-level specialization while suppressing interference from unrelated branches.

\noindent \textbf{Generalization to new clients.}  
For a new client $u$, we assign a cluster using the same subspace metric as in LoRA-Subspace Adaptive Clustering. Specifically, we extract a lightweight probe basis $\mathbf{B}_u$ by running a few local gradient steps, compute its top-$r$ left singular vectors $\mathbf{U}_u$, and determine
$
j^\star(u) \;=\; 
\arg\max_j \;
\operatorname{mean}\!\Big(\cos^2 \Theta\!\big(\mathbf{U}_u,\, \mathbf{U}_{c,j}\big)\Big),
$
where $\mathbf{U}_{c,j}$ denotes the subspace of cluster $j$, and $\Theta(\cdot,\cdot)$ represents the set of principal angles. This procedure enables immediate inference through root and cluster tiers, while the leaf adapter can be instantiated and refined online.

\section{Theoretical Guarantees}
\vspace{-1mm}
\noindent \textbf{Notation.}
We consider \(N\) clients partitioned into \(K\) clusters, as in \cref{eq:three_levels}.
Client \(i\) has data distribution \(D_i\) and dataset \(\mathcal D_i \sim D_i^{\,n_i}\).
Let \(D\) denote the global distribution induced by \(\{D_i\}_{i=1}^N\) and \(C_j\) the cluster distribution induced by \(\{D_i: i\in\mathcal C_j\}\).
We independently draw global samples \(S\sim D^{\,m}\), cluster-level samples \(S_j\sim C_j^{\,m_j}\), and client-level samples \(\mathcal D_i\sim D_i^{\,n_i}\).
For any function class $\Fcal$ and distributions $Q_1,Q_2$,
$
\disc_{\Fcal}(Q_1,Q_2):=\sup_{f\in\Fcal}\big|L_{Q_1}(f)-L_{Q_2}(f)\big|.
$
Let \(\Hcal\) be the hypothesis class consistent with the HiLoRA parameterization \(\Phi\) in \cref{eq:pfed_obj}.
Denote by \(\Gcal_r,\Gcal_c,\Gcal_\ell\) the root, cluster, and leaf prediction classes, 
and let the corresponding loss classes be
\(\Fcal_r := \{\,L(h):h\in\Gcal_r\}\), 
\(\Fcal_c := \{\,L(h):h\in\Gcal_c\}\), 
\(\Fcal_\ell := \{\,L(h):h\in\Gcal_\ell\}\). Let \(\mathcal Y=\{1,\dots,C\}\) and the predictor output lie in \(\mathbb{R}^{C}\).
The loss \(L:\mathbb{R}^{C}\times\mathcal Y\to[0,1]\) is bounded.
\vspace{-2mm}
\begin{assumption}[Orthogonal LoRA Subspaces]\label{ass:orth}
For each client $i$ with cluster $j(i)$, there exist subspaces
$U_r,\,\{U_c^{(j)}\}_{j=1}^K,\,\{U_\ell^{(i)}\}_{i=1}^N \subseteq \mathbb R^{p\times q}$
such that, w.r.t.\ the Frobenius inner product,
$U_c^{(j(i))}\perp U_r$ and $U_\ell^{(i)}\perp (U_r\oplus U_c^{(j(i))})$.
Tier-wise LoRA updates satisfy
$\Delta W_r\in U_r$, $\Delta W_c^{(j(i))}\in U_c^{(j(i))}$, $\Delta W_\ell^{(i)}\in U_\ell^{(i)}$.
We define the restricted loss classes:
\[
\Fcal_c^{\perp(j)} := \{(x,y)\mapsto L(h^{(r)}(x;W_0+\Delta W_c),y)\},
\]
\[
\Fcal_\ell^{\perp(i)} := \{(x,y)\mapsto L(h^{(r,c)}(x;W_0+\Delta W_\ell),y)\}.
\]
\end{assumption}

\begin{tcolorbox}[graytheorem]
\begin{theorem}[HiLoRA Excess-Risk Generalization Bound]\label{thm:main}
Under Assumption \ref{ass:orth}, for any client \(i\) and \(\delta\in(0,1)\), with a probability of at least \(1-3\delta\),
\begin{align}
\label{eq:leaf-final}
&L_{D_i}\!\big(h^{(r,c,\ell)}\big)-\inf_{h\in\Hcal} L_{D_i}(h)\nonumber\\
\le&\big(\hat L_{S}(h^{(r)})-\inf_{h\in\Hcal}\hat L_{S}(h)\big)
+ 4\,\Rehat_{S}(\Fcal)
+  6\sqrt{\tfrac{\log(2/\delta)}{2m}}
\nonumber\\&+  2\,\disc_{\Fcal}(D_i,D)
+\big(\hat L_{S_{j(i)}}(h^{(r,c)})-\hat L_{S_{j(i)}}(h^{(r)})\big)
\nonumber\\&+  4\,\Rehat_{S_{j(i)}}\!\big(\Fcal_c^{\perp(j(i))}\big)
+ 2\,\disc_{\Fcal_c^{\perp(j(i))}}(D_i,C_{j(i)})\nonumber\\
&+ 6\sqrt{\tfrac{\log(2/\delta)}{2m_{j(i)}}}+\big(\hat L_{\mathcal D_i}(h^{(r,c,\ell)})-\hat L_{\mathcal D_i}(h^{(r,c)})\big)
\nonumber\\&+ 4\,\Rehat_{\mathcal D_i}\!\big(\Fcal_\ell^{\perp(i)}\big)+ 6\sqrt{\tfrac{\log(2/\delta)}{2n_i}}.
\end{align}
\end{theorem}
\end{tcolorbox}
\noindent Theorem~\ref{thm:main} is proved using Lemmas~1--2, with detailed proofs deferred to Appendix~B. We partition the bound in \eqref{eq:leaf-final} into three components. First, the Generalization (\(\mathbf{GE}\)) terms comprise the empirical Rademacher complexity \(4\Rehat_{S}(\Fcal)\), \(4\Rehat_{S_{j(i)}}(\Fcal_c^{\perp(j(i))})\), and \(4\Rehat_{\mathcal D_i}(\Fcal_\ell^{\perp(i)})\), together with their concentration terms \(6\sqrt{\tfrac{\log(2/\delta)}{2m}}\), \(6\sqrt{\tfrac{\log(2/\delta)}{2m_{j(i)}}}\), and \(6\sqrt{\tfrac{\log(2/\delta)}{2n_i}}\). Second, the Distribution Shift (\(\mathbf{DS}\)) terms consist of \(2\,\disc_{\Fcal}(D_i,D)\) and \(2\,\disc_{\Fcal_c^{\perp(j(i))}}(D_i,C_{j(i)})\). Third, the Empirical Optimization (\(\mathbf{EO}\)) terms are \(\big(\hat L_{S}(h^{(r)})-\inf_{h\in\Hcal}\hat L_{S}(h)\big)\ (\ge 0)\), \(\big(\hat L_{S_{j(i)}}(h^{(r,c)})-\hat L_{S_{j(i)}}(h^{(r)})\big)\ (\le 0)\), and \(\big(\hat L_{\mathcal D_i}(h^{(r,c,\ell)})-\hat L_{\mathcal D_i}(h^{(r,c)})\big)\ (\le 0)\). 

\begin{corollary}
Under Assumption \ref{ass:orth}, for any client \(i\) and \(\delta\in(0,1)\), with a probability of at least \(1-3\delta\),
\begin{align}
L_{D_i}\!\big(h^{(r,c,\ell)}\big)
-\inf_{h\in\Hcal}L_{D_i}(h)
\;\le\;
\mathbf{GE}
+\mathbf{DS}
+\mathbf{EO}.
\end{align}
\end{corollary}

\begin{table*}[h] 
\caption{\textbf{Performance  analysis on CIFAR-100.} We evaluate across statistical heterogeneity patterns: \emph{(i) Client-level Personalization}: reporting the mean (Avg) and worst-case (10\%) accuracies $\pm$ std across all clients. \emph{(ii) Unseen-Client Adaptation}: accuracy on unseen clients’ test distributions. Best results are in \textbf{bold}; second-best are \underline{underlined}.} 
\label{tab:main_cvpr_80} 
\centering 
\small 
\setlength{\tabcolsep}{3.20pt} \renewcommand{\arraystretch}{1.19} 
  \resizebox{\textwidth}{!}{\begin{tabular}{l cc cc cc ccc} \toprule \multirow{3}{*}{\textbf{Method}} & \multicolumn{6}{c}{\textbf{(i) Personalization Accuracy (↑)}} & \multicolumn{3}{c}{\textbf{(ii) Unseen-Client Adaptation (↑)}}\\ \cmidrule(lr){2-7}\cmidrule(lr){8-10} & \multicolumn{2}{c}{\textbf{GL–Dir(0.3)}} & \multicolumn{2}{c}{\textbf{SC–Dir(3)}} & \multicolumn{2}{c}{\textbf{Patho(10)}} & \textbf{GL–Dir(0.3)} & \textbf{SC–Dir(0.3)} & \textbf{Patho(10)} \\ & \textbf{Mean Acc.} & \textbf{Worst Acc.} & \textbf{Mean Acc.} & \textbf{Worst Acc.} & \textbf{Mean Acc.} & \textbf{Worst Acc.} & \textbf{Test Acc.} & \textbf{Test Acc.} & \textbf{Test Acc.} \\ \midrule

Local-LoRA  & $0.691\scriptstyle \pm 0.06$ & $0.579\scriptstyle \pm0.03$ & $0.867 \scriptstyle \pm 0.11$ & $0.601\scriptstyle \pm 0.10$ & $0.844\scriptstyle \pm0.12$ & $0.577\scriptstyle \pm0.09$  & $0.521\scriptstyle \pm0.09$  & $0.639\scriptstyle \pm0.17$ & $0.756\scriptstyle \pm0.09$ \\

FedIT  & $0.772\scriptstyle \pm 0.05$ & $0.711 \scriptstyle \pm 0.04$ & $0.578\scriptstyle \pm0.16$ & $0.315 \scriptstyle \pm 0.03$ & $0.869\scriptstyle \pm0.07$ & $0.766\scriptstyle \pm0.07$ & $0.753\scriptstyle \pm 0.04$ & $0.687\scriptstyle \pm 0.16$ & $0.848\scriptstyle \pm0.17$ \\
FlexLoRA   & $\underline{0.818\scriptstyle \pm 0.06}$ & $0.695\scriptstyle \pm0.04$ & $0.718\scriptstyle \pm 0.20$ & $0.301\scriptstyle \pm 0.15$ & $0.862\scriptstyle \pm0.06$ & $0.775\scriptstyle \pm0.06$ & $0.808\scriptstyle \pm0.05$ & $0.719\scriptstyle \pm0.17$  & $0.834\scriptstyle \pm 0.19$ \\

FedSA-LoRA   & $0.786 \scriptstyle\pm 0.05$ & $0.693 \scriptstyle\pm 0.02$& $0.734\scriptstyle \pm 0.12$ & $0.462\scriptstyle \pm0.08$ & $0.860\scriptstyle\pm0.09$ & $0.638\scriptstyle \pm0.11$  & $0.813\scriptstyle\pm 0.06$ & $0.765\scriptstyle \pm 0.10$ & $0.846\scriptstyle\pm0.09$ \\

FDLoRA     & $0.802\scriptstyle \pm0.03$ & $0.731\scriptstyle \pm0.04$ & $0.828\scriptstyle \pm0.14$ & $0.673\scriptstyle \pm0.12$ & $0.871\scriptstyle \pm0.10$ & $0.774\scriptstyle \pm0.08$ & $0.754 \scriptstyle \pm 0.05$  & $0.701 \scriptstyle \pm 0.15$ & $0.813\scriptstyle \pm0.13$ \\
FedDPA-F   & $0.811\scriptstyle \pm 0.06$ & $\underline{0.743\scriptstyle \pm0.04}$& $0.841\scriptstyle \pm 0.13$ & $0.574\scriptstyle \pm 0.09$ & $0.905\scriptstyle \pm0.08$ & $0.701\scriptstyle \pm0.09$  & $\underline{0.821 \scriptstyle \pm 0.03}$ & $0.686 \scriptstyle \pm 0.15$ & $0.854\scriptstyle \pm0.16$ \\
FedDPA-T  & $0.803\scriptstyle \pm0.04$ & $0.726\scriptstyle \pm0.02$ & $0.895\scriptstyle \pm0.06$ & $0.745\scriptstyle \pm0.07$  & $0.928\scriptstyle \pm0.05$ & $\underline{0.818\scriptstyle \pm0.04}$ & $0.810 \scriptstyle \pm 0.04$ & $0.738 \scriptstyle \pm 0.13$  & $\underline{0.879\scriptstyle \pm0.10}$ \\
PF2LoRA    & $0.793\scriptstyle \pm 0.03$ & $0.714\scriptstyle \pm 0.02$ & $0.873\scriptstyle \pm 0.08$ & $0.726 \scriptstyle \pm 0.10$ & $0.916\scriptstyle \pm0.05$ & $0.794\scriptstyle \pm0.05$  & $0.802\scriptstyle \pm 0.03$ & $\underline{0.788\scriptstyle \pm 0.12}$ &  $0.860\scriptstyle \pm0.08$ \\
FedALT    & $0.809\scriptstyle \pm 0.04$ & $0.695\scriptstyle \pm0.04$ & $\underline{0.912\scriptstyle \pm 0.06}$ & $\underline{0.763 \scriptstyle \pm 0.06}$ & $\underline{0.929\scriptstyle \pm0.06}$ & $0.801\scriptstyle \pm0.05$ & $0.756\scriptstyle \pm0.07$ & $0.774\scriptstyle \pm 0.14$ & $0.763\scriptstyle \pm0.11$  \\
\specialrule{0.3pt}{1.4pt}{0.2pt}
\rowcolor{gray!20}\bfseries
\textbf{HiLoRA} &
  \textbf{0.846±0.04} & \textbf{0.763±0.02} &
   \textbf{0.934±0.06} & \textbf{0.791±0.06} &
  \textbf{0.941±0.05} & \textbf{0.823±0.03} & \textbf{0.841±0.04} &
  \textbf{0.859±0.11} & \textbf{0.940±0.04} \\
\specialrule{0.6pt}{0pt}{0pt}
\end{tabular}
}
\end{table*}

\begin{table*}[h]
\caption{\textbf{Comparative analysis on DomainNet.}
    We evaluate: \emph{(i) Client-level Personalization}, reporting the mean (Avg) and worst-case (10\%) accuracies $\pm$ std across all clients;
    \emph{(ii) Domain-wise accuracy}, averaging  client accuracies within each domain (C/I/P/Q/R/S);
    \emph{(iii) Unseen-Client Adaptation}, accuracy on test distributions of unseen clients.
    Best results are in \textbf{bold}; second-best are \underline{underlined}.}
  \label{tab:main_domainnet}
  \centering
  \setlength{\tabcolsep}{3.25pt}
  \renewcommand{\arraystretch}{1.19}
  \resizebox{\textwidth}{!}{%
  \begin{tabular}{l cc cccccc c}
    \toprule
    \multirow{2}{*}{\textbf{Method}} &
      \multicolumn{2}{c}{\textbf{(i) Client-level (↑)}} &
      \multicolumn{6}{c}{\textbf{(ii) Domain-wise Accuracy (↑)}} &
      \textbf{(iii) Unseen (↑)} \\
    \cmidrule(lr){2-3}\cmidrule(lr){4-9}\cmidrule(lr){10-10}
     & \textbf{Mean Acc.} & \textbf{Worst Acc.}
     & \textbf{C} & \textbf{I} & \textbf{P} & \textbf{Q} & \textbf{R} & \textbf{S}
     & \textbf{Test Acc.} \\
    \midrule
    Local-LoRA  & $0.755\scriptstyle \pm 0.17$ & $0.374\scriptstyle \pm 0.11$
                & $0.716\scriptstyle \pm 0.07$ & $0.428\scriptstyle \pm 0.12$ & $0.850\scriptstyle \pm 0.06$ 
                & $0.748\scriptstyle \pm 0.06$ & $0.954\scriptstyle \pm 0.02$ & $0.794\scriptstyle \pm 0.05$
                & $0.665\scriptstyle \pm 0.21$ \\
    FedIT    & $0.822\scriptstyle \pm 0.15$ & $0.494\scriptstyle \pm 0.05$
                & $0.886\scriptstyle \pm 0.09$ & $0.539\scriptstyle \pm 0.08$ & $0.898\scriptstyle \pm 0.06$
                & $0.762\scriptstyle \pm 0.07$ & $0.966\scriptstyle \pm 0.02$ & $0.854\scriptstyle \pm 0.03$
                & $0.813\scriptstyle \pm 0.14$ \\
    FlexLoRA     & $0.839\scriptstyle \pm 0.14$ & $0.526\scriptstyle \pm 0.07$
               & $\underline{0.887\scriptstyle \pm 0.05}$ & $0.567\scriptstyle \pm 0.09$ & $\textbf{0.926±0.06}$
              & $0.779\scriptstyle \pm 0.05$ & $0.975\scriptstyle \pm 0.02$ & $\textbf{0.874±0.04}$
             & $\underline{0.840\scriptstyle \pm 0.11}$ \\
    FedSA-LoRA  & $0.814\scriptstyle \pm 0.13$ & $0.532\scriptstyle \pm 0.05$
                & $0.850\scriptstyle \pm 0.07$ & $0.568\scriptstyle \pm 0.07$ & $0.867\scriptstyle \pm 0.06$
                & $0.810\scriptstyle \pm 0.07$ & $0.971\scriptstyle \pm 0.02$ & $0.792\scriptstyle \pm 0.06$
                & $0.797\scriptstyle \pm 0.17$ \\
    FDLoRA      & $0.822\scriptstyle \pm 0.13$ & $0.535\scriptstyle \pm 0.05$
                & $0.860\scriptstyle \pm 0.07$ & $0.562\scriptstyle \pm 0.07$ & $0.882\scriptstyle \pm 0.05$
                & $0.822\scriptstyle \pm 0.06$ & $0.974\scriptstyle \pm 0.02$ & $0.800\scriptstyle \pm 0.05$
                & $0.804\scriptstyle \pm 0.17$ \\
    FedDPA-F    & $\underline{0.860\scriptstyle \pm 0.12}$ & $\underline{0.583\scriptstyle \pm 0.04}$
                & $0.879\scriptstyle \pm 0.07$ & $\underline{0.610\scriptstyle \pm 0.06}$ & $0.922\scriptstyle \pm 0.04$
                & $\underline{0.877\scriptstyle \pm 0.05}$ & $\underline{0.988\scriptstyle \pm 0.02}$ & $0.848\scriptstyle \pm 0.05$
                & $0.831\scriptstyle \pm 0.16$ \\
    FedDPA-T    & $0.831\scriptstyle \pm 0.13$ & $0.529\scriptstyle \pm 0.02$
                & $0.875\scriptstyle \pm 0.06$ & $0.554\scriptstyle \pm 0.05$ & $0.897\scriptstyle \pm 0.06$
                & $0.830\scriptstyle \pm 0.04$ & $0.976\scriptstyle \pm 0.01$ & $0.819\scriptstyle \pm 0.04$
                & $0.835\scriptstyle \pm 0.13$ \\
    PF2LoRA     & $0.820\scriptstyle \pm 0.13$ & $0.531\scriptstyle \pm 0.05$
                & $0.858\scriptstyle \pm 0.07$ & $0.560\scriptstyle \pm 0.07$ & $0.880\scriptstyle \pm 0.05$
                & $0.820\scriptstyle \pm 0.06$ & $0.973\scriptstyle \pm 0.02$ & $0.799\scriptstyle \pm 0.05$
                & $0.801\scriptstyle \pm 0.18$ \\
    FedALT      & $0.796\scriptstyle \pm 0.15$ & $0.440\scriptstyle \pm 0.04$
                & $0.804\scriptstyle \pm 0.07$
                & $0.482\scriptstyle \pm 0.08$ & $0.849\scriptstyle \pm 0.05$ & $0.741\scriptstyle \pm 0.08$
                & $0.953\scriptstyle \pm 0.02$ & $0.797\scriptstyle \pm 0.04$ & $0.743\scriptstyle \pm 0.21$ \\
     \specialrule{0.3pt}{1.4pt}{0.2pt}
        \rowcolor{gray!20}
    \textbf{HiLoRA}
                & \textbf{0.877±0.11}& \textbf{0.589±0.02}
                & \textbf{0.921±0.06} & \textbf{0.624±0.06}
                & \underline{$0.924\scriptstyle \pm 0.03$} & \textbf{0.899±0.05}
                & \textbf{0.989±0.01} & \underline{$0.872\scriptstyle \pm 0.06$}
                & \textbf{0.861±0.12} \\
    \specialrule{0.6pt}{0pt}{0pt}
    \vspace{-8mm}
  \end{tabular}%
  }
\end{table*}
\noindent\textbf{Remark 1.}
HiLoRA tightens the GE terms via sample amplification and orthogonality.
The root depends on \(m=\sum_{u=1}^N n_u\) and the cluster-level bound on \(m_{j(i)}=\sum_{t\in\mathcal C_{j(i)}} n_t\), yielding $O(m^{-1/2})$ and $O(m_{j(i)}^{-1/2})$ rates. Meanwhile, $\Fcal_c^{\perp(j(i))}\subseteq\Fcal_c$ and $\Fcal_\ell^{\perp(i)}\subseteq\Fcal_\ell$, orthogonality narrows the classes further reducing their  complexities.

\noindent\textbf{Remark 2.}
The DS terms are reduced via cluster-aware grouping by assigning client $i$ to $j(i)$, making $D_i$ closer to $C_{j(i)}$, thereby lowering $\,\disc_{\Fcal_c^{\perp(j(i))}}(D_i,C_{j(i)})$.

\noindent\textbf{Remark 3.}
With orthogonality ($U_c^{(j(i))}\!\perp\! U_r$, $U_\ell^{(i)}\!\perp\!(U_r\oplus U_c^{(j(i))})$), the cluster and leaf empirical terms are nonincreasing on their own data. Defining gains
$G_c(i):=\hat L_{S_{j(i)}}(h^{(r)})-\hat L_{S_{j(i)}}(h^{(r,c)})\ge 0$ and
$G_\ell(i):=\hat L_{\mathcal D_i}(h^{(r,c)})-\hat L_{\mathcal D_i}(h^{(r,c,\ell)})\ge 0$,
these appear in the bound as ($-G_c(i)-G_\ell(i)$), tightening the bound numerically.

\section{Experiments}
\subsection{Experimental Setup}
\noindent\textbf{Model and Datasets.}
We adopt the ViT-Base pretrained on ImageNet-21K as the backbone \cite{dosovitskiy2021an} and insert LoRA adapters into the query and value projections of each attention layer. Experiments are conducted on CIFAR-100 and DomainNet. For CIFAR-100, we instantiate 100 clients with three label-skewed non-IID partitions.
\emph{GL–Dir($\alpha$)} samples client label priors from a Dirichlet distribution over all classes \cite{gao2022feddc}, using $\alpha=0.3$ for strong skew.
\emph{SC–Dir($\alpha$)} extends Dirichlet partitioning to superclasses \cite{kim2023protofl,zhu2023confidence,zhang2025pfllib}; each client’s superclass prior is drawn from a Dirichlet with $\alpha=3$, producing data concentrated on one to two superclasses.
\emph{Patho(10)} follows the pathological non-IID setup with 10 classes per client \cite{mcmahan2017communication}. For DomainNet, we construct 90 clients across six domains, each representing a distinct \emph{domain skew}, where intra-domain label sampling follows a Dirichlet prior distribution with $\alpha$ set to 0.6 \cite{su2024federated,yan2024local}.  Consistent with \cite{yang2023efficient,wang2024aggregation}, we use the ten most frequent classes to build a sub-dataset for our experiments.

\noindent\textbf{Baselines.}
We compare HiLoRA with nine LoRA-based federated learning baselines. Local-LoRA fine-tunes local data without aggregation. Single-branch federated adapters include FedIT \cite{zhang2024towards}, which directly aggregates client $\mathbf{B}$ and $\mathbf{A}$ modules, and FlexLoRA \cite{bai2024federated}, which aggregates in the product space $\mathbf{B}\mathbf{A}$. Dual-LoRA personalization methods include FedSA-LoRA \cite{guo2025selective}, which uses a shared basis $\mathbf A$ and local coefficients $\mathbf B$; FDLoRA \cite{qi2024fdlora}, which alternates updates of shared and client adapters; FedDPA-F and FedDPA-T \cite{long2024dual}, which adopt sequential and alternating training, respectively; and PF2LoRA \cite{hao2025personalized}, which combines a global and a personal branch with automatic rank selection. FedALT \cite{bian2025fedalt} mixes an individual adapter with a rest-of-the-world adapter through an adaptive gating network.


\noindent\textbf{Evaluation and Implementation Details.}  
\textit{(1) Personalization.} Each client is evaluated on its own test set. We report the mean accuracy across clients and the 10th-percentile accuracy to assess tail robustness\cite{marfoq2022personalized,marfoq2021federated}.  
\textit{(2) Generalization.} We hold out 20\% of clients as unseen for evaluating generalization~\cite{marfoq2022personalized,wu2023personalized}. At test time, each unseen client performs 5 local epochs of fine-tuning on the resulting model before evaluation on its test set \cite{fallah2020personalized,oh2022fedbabu}. 
We train for 50 global rounds, using 1 local epoch for CIFAR-100 and 2 local epochs for DomainNet.  
To ensure a fair comparison, HiLoRA maintains an equal total training budget, with  
$T_{\text{root}} + T_{\text{cluster}} + T_{\text{leaf}} = 50$,  
to match that of all baselines. Detailed implementation and training settings are described in Appendix D.

\begin{figure}[t]
  \centering
  \subfloat[\textbf{Patho(10)}]{
    \includegraphics[width=0.46\linewidth]{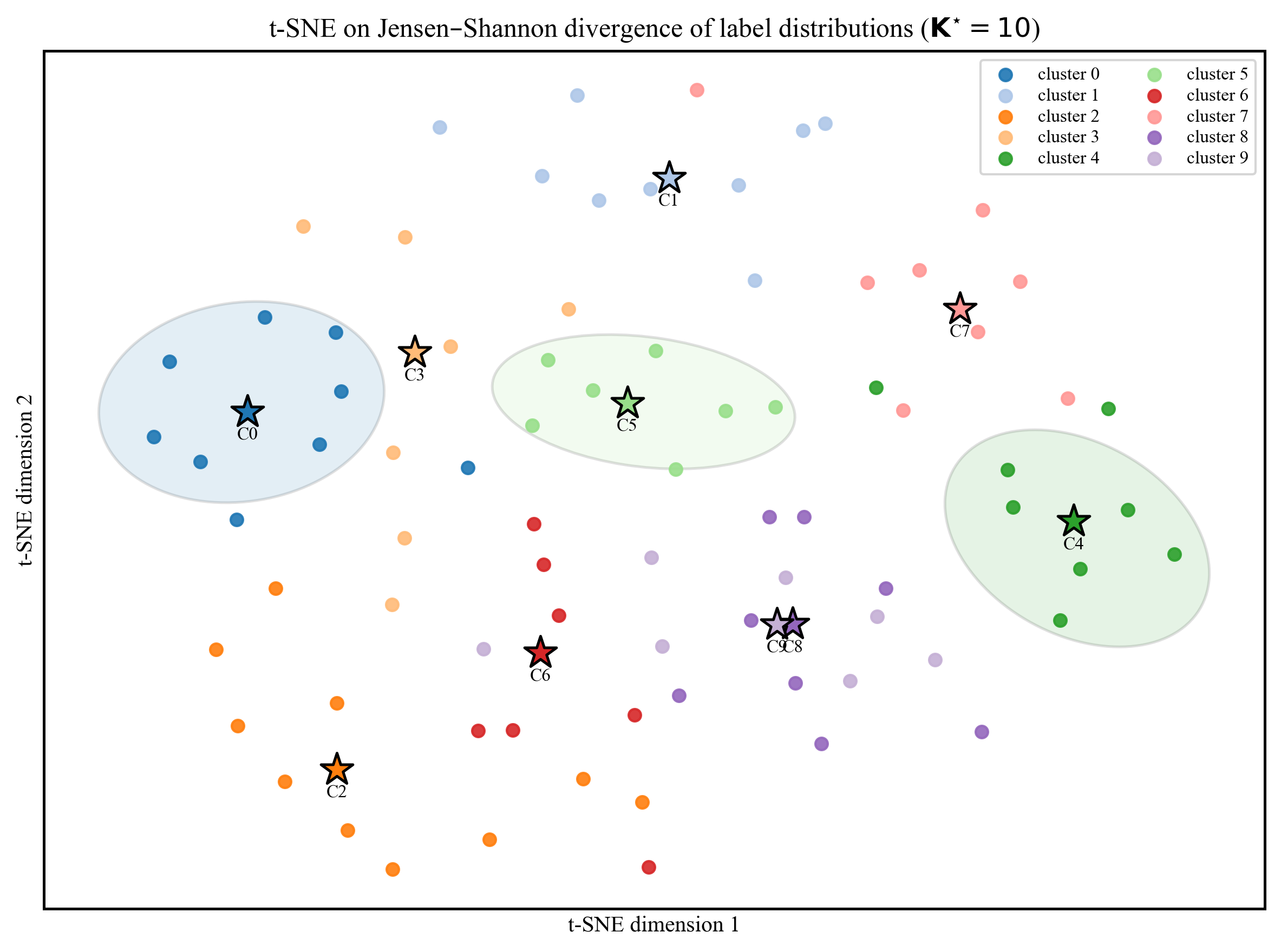}
    \label{fig:gsac-path10}
  }\hfill
  \subfloat[\textbf{SC–Dir(3)}]{
    \includegraphics[width=0.46\linewidth]{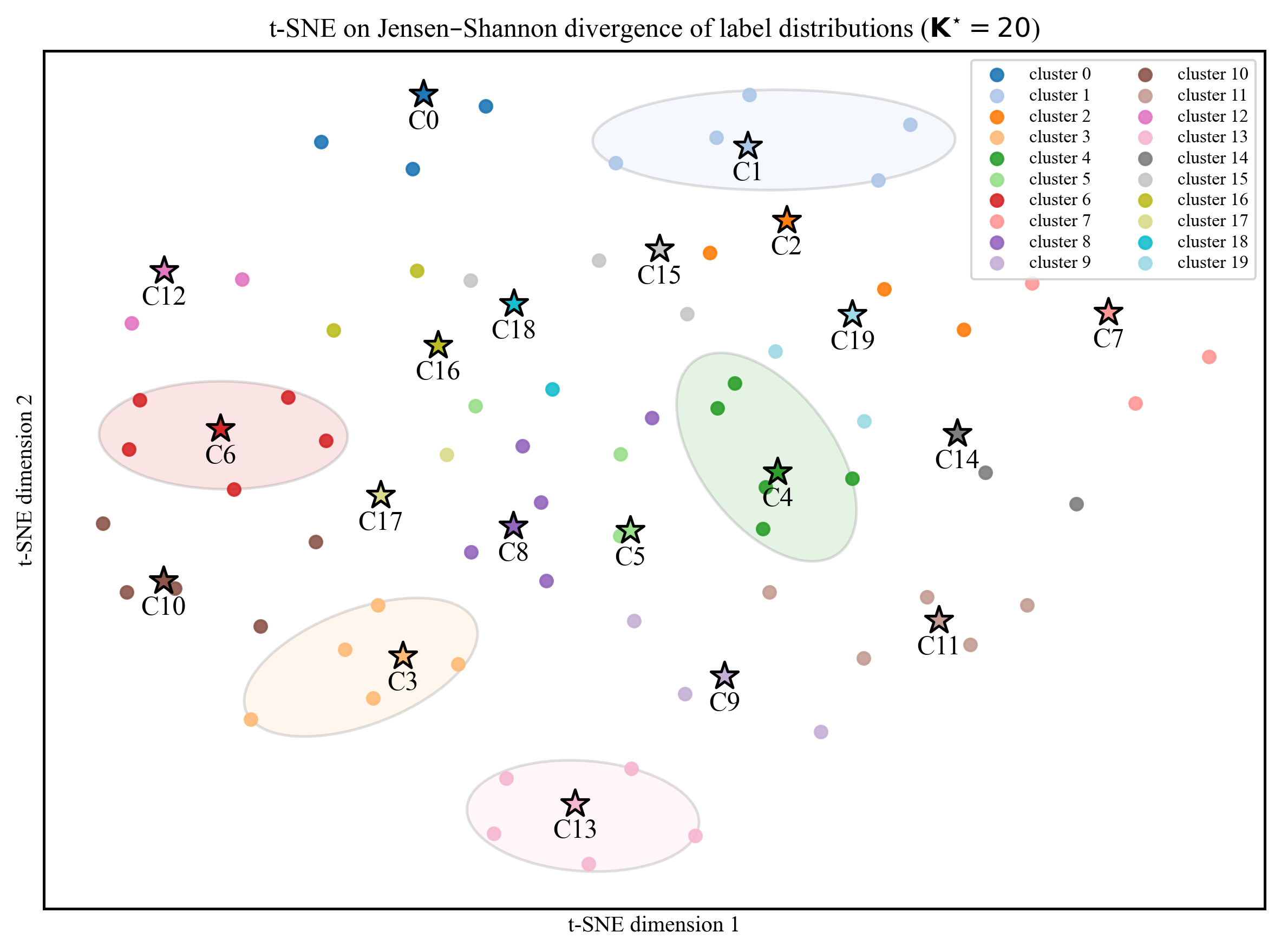}
    \label{fig:gsac-exdir3}
  }
    \caption{\textbf{t-SNE of clustering results on CIFAR-100.}
    Clients are visualized by Jensen–Shannon distances of their label distributions: (a) Patho-10, $K^{\star}{=}10$; (b) SC–Dir($\alpha{=}3$), $K^{\star}{=}20$.}
  \label{fig:gsac-tsne}
  \vspace{-5mm} 
\end{figure}

\subsection{Performance Evaluation}
\subsubsection{Label-Heterogeneous Federated Setting}
\noindent \textbf{Overall Results on CIFAR-100.} Table~\ref{tab:main_cvpr_80} summarizes the overall performance of HiLoRA on the CIFAR-100 dataset under different label heterogeneity settings. For \emph{personalization}, HiLoRA consistently achieves the best mean and worst-case accuracies across all partitions, surpassing prior methods such as FedALT and FedDPA-T. For instance, under the SC–Dir(3) setting, HiLoRA improves mean accuracy from 0.912 to 0.934 and worst-case accuracy from 0.763 to 0.791 over the second-best baseline, demonstrating the advantage of hierarchical three-tier adaptation over dual-branch and other flat LoRA designs.  \par\vspace{-0.1em}

For generalization ability, HiLoRA delivers leading performance across all configurations \emph{at a comparable training cost to other methods}. This advantage arises from adaptive clustering, which groups clients with similar data distributions to share knowledge and form stable cluster-level representations. At test time, an unseen client is routed to its nearest cluster using LoRA subspace similarity and reuses the corresponding adapter, enabling faster and more effective adaptation to new clients.

\noindent \textbf{Impact of Label Heterogeneity Levels.}
Results in Table~\ref{tab:main_cvpr_80} indicate that HiLoRA exhibits stable performance across varying degrees of label heterogeneity, with both personalization and generalization remaining robust compared to other approaches.  This performance is attributed to HiLoRA's hierarchical adaptation mechanism, which effectively suppresses cross-distribution interference, enabling the model to maintain consistent learning performance across diverse label-distribution patterns.
\begin{figure}[t]
    \centering
    \begin{subfigure}[t]{0.49\columnwidth}
        \centering
        \includegraphics[width=\linewidth]{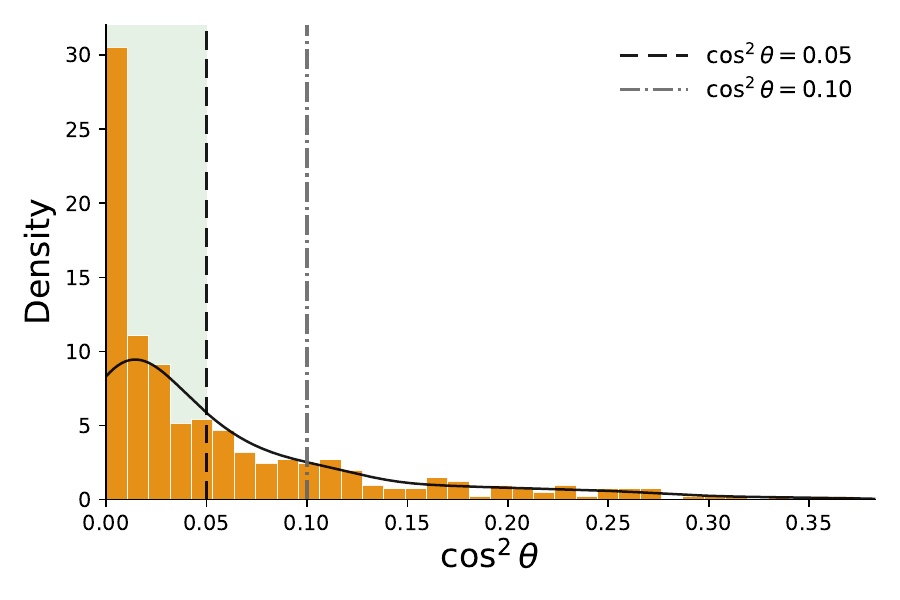}
        \caption{Root vs Leaf.}
        \label{fig:pa-root-leaf}
    \end{subfigure}
    \hfill
    \begin{subfigure}[t]{0.49\columnwidth}
        \centering
        \includegraphics[width=\linewidth]{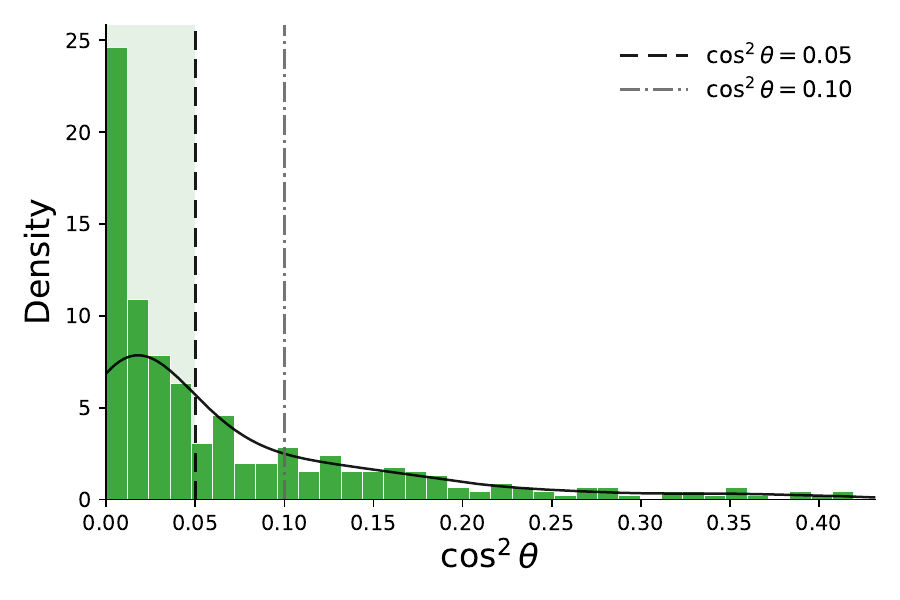}
        \caption{Cluster vs Leaf.}
        \label{fig:pa-cluster-leaf}
    \end{subfigure}
    \vspace{-1mm}
    \caption{\textbf{Principal-angle distributions between tiers in HiLoRA.}
    Computed from the column spaces of the LoRA-$\mathbf{B}$ matrices. Lower $\cos^2\theta$ indicates stronger subspace orthogonality. }
    \label{fig:hilora-orth-singlecol}
    \vspace{-2.4mm}
\end{figure}

\begin{table}[h]
\centering
\caption{\textbf{HiLoRA multi-level gain comparison across Root, Cluster, and Leaf stages} on CIFAR-100 and DomainNet, reporting Mean±Std and accuracy gains relative to Root and Cluster.}
\vspace{-1.5mm}
\label{tab:dataset_stage_gains}
\setlength{\tabcolsep}{5pt}
\renewcommand{\arraystretch}{1.4}
\resizebox{\columnwidth}{!}{%
\begin{tabular}{l rr rr}
\toprule
\textbf{Stage} & \multicolumn{2}{c}{\textbf{CIFAR-100}} & \multicolumn{2}{c}{\textbf{DomainNet}} \\
\cmidrule(lr){2-3}\cmidrule(lr){4-5}
& \textbf{Mean} $\pm \textbf{std}$ & \textbf{Gains (R/C) \%} & \textbf{Mean} $\pm \textbf{std}$ & \textbf{Gains (R/C) \%} \\
\midrule
Root        & 0.663 $\pm$ 0.18        &  ----/----        & 0.815 $\pm$ 0.15 & ----/---- \\
\rowcolor{gray!8}
+ Cluster   & 0.889 $\pm$ 0.10       & +22.6\% / ----          & 0.864 $\pm$ 0.13 & +4.9\% / ---- \\
\rowcolor{gray!16}
+ Leaf      & \textbf{0.934} $\pm$ \textbf{0.06}         & \textbf{+27.1\% / +4.5\%}           & \textbf{0.877 $\pm$ 0.11} & \textbf{+6.2\% / +1.3\%} \\    \specialrule{0.6pt}{0pt}{0pt}
\emph{Second-best} 
& \multicolumn{2}{c}{\underline{0.912$\pm$0.06}} 
& \multicolumn{2}{c}{\underline{0.860$\pm$0.12}} \\
\bottomrule
\end{tabular}}
\vspace{-4.9mm} 
\end{table}
\vspace{-1.0mm} 
\subsubsection{Domain-Heterogeneous Federated Setting}

\noindent \textbf{Overall Results on DomainNet.} As shown in Table \ref{tab:main_domainnet}, HiLoRA consistently surpasses all baselines on DomainNet in both mean and tail-client accuracy, demonstrating strong robustness under domain distribution shifts. Compared with the next-best baselines, FedDPA-F and FlexLoRA, HiLoRA improves mean accuracy by about 1.5\%–2.0\% while maintaining the highest generalization performance on unseen clients. These results highlight the effectiveness of hierarchical decomposition in transferring shared representations across domains while preserving client-level adaptability.

\noindent \textbf{Domain-Specific Personalization.} 
The results in Table~\ref{tab:main_domainnet} indicate that HiLoRA attains strong performance across all domains, especially in challenging ones such as illustration (I) and sketch (S). This indicates that the hierarchical design with adaptive clustering yields domain-specific adapters and suppresses cross-domain interference, enabling refined and disentangled adaptation within each domain.

\vspace{-1.2mm}
\subsection{Ablation and Further Analysis}
In this section, we present detailed analyses of each module and provide further analysis in Appendix E.

\noindent\textbf{Progressive Layer Gains.} As shown in Table \ref{tab:dataset_stage_gains}, the model’s performance steadily improves with each additional tier. On CIFAR-100, the accuracy rises from 0.663 at the Root stage to 0.889 with the Cluster layer, yielding an absolute gain of 22.6\%. Incorporating the Leaf layer further improves accuracy to 0.934, which corresponds to a 27.1\% increase over Root and a 4.5\% increase over Cluster. Meanwhile, the standard deviation consistently decreases from 0.18 to 0.06, indicating more uniform performance across clients. A similar trend appears on DomainNet, where accuracy improves at each tier while variance declines.

\begin{figure}[t]
  \centering
  \begin{subfigure}{0.48\linewidth}
    \centering
    \includegraphics[width=\linewidth]{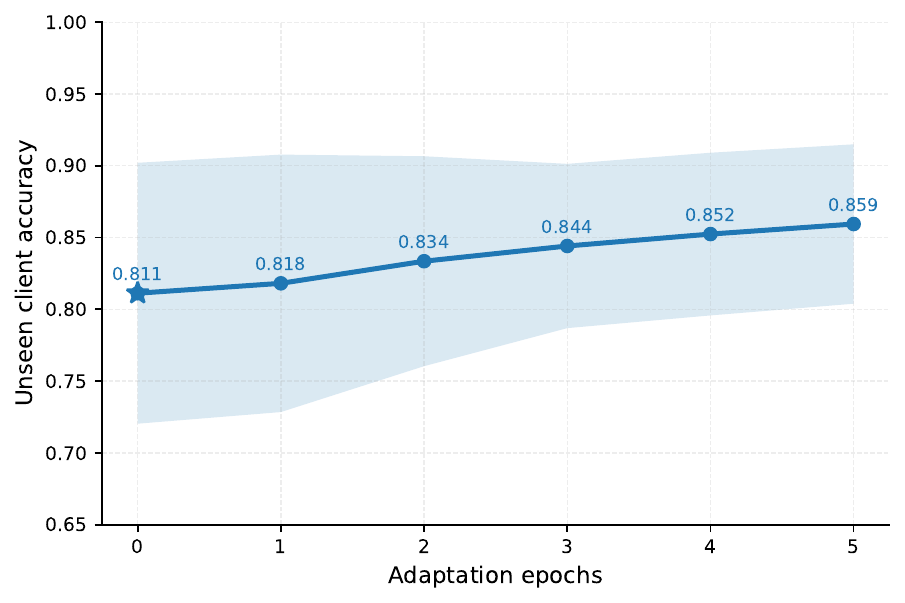}
    \subcaption{CIFAR-100}
  \end{subfigure}
  \hfill
  \begin{subfigure}{0.48\linewidth}
    \centering
    \includegraphics[width=\linewidth]{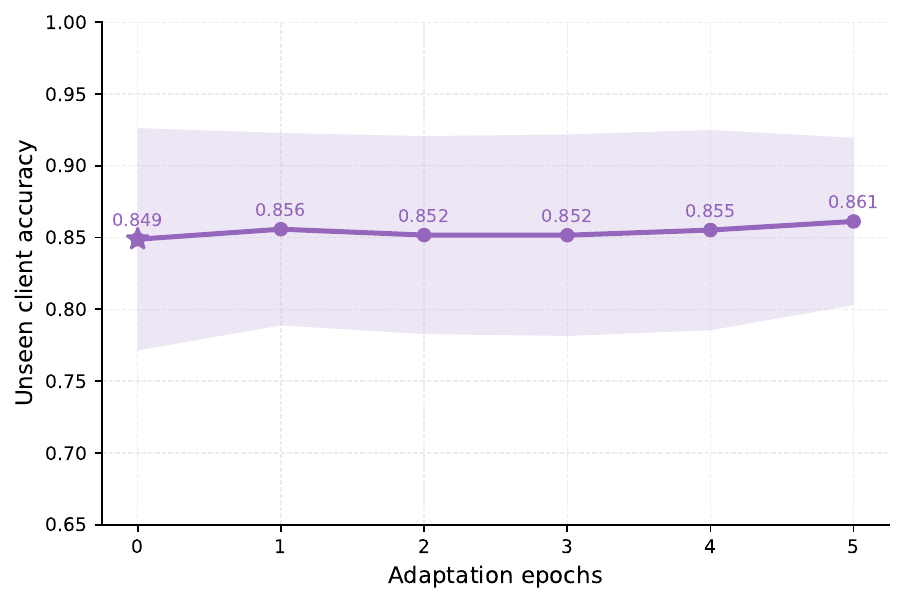}
    \subcaption{DomainNet}
  \end{subfigure}
  \caption{Unseen-client accuracy (\emph{mean} ± \emph{std}) across adaptation epochs on \textbf{(a) CIFAR-100} and \textbf{(b) DomainNet}. The starred point indicates the Root+Cluster initialization, while subsequent epochs correspond to leaf-level LoRA adaptation.}
  \label{fig:adapt_epochs}
  \vspace{-6pt} 
\end{figure}

\begin{table}[t]
\centering
\caption{\textbf{Ablation study of HiLoRA components.}
We report personalized accuracy (Per.) and unseen-client adaptation (Gen.) on CIFAR-100 and DomainNet.}
\vspace{-1.5mm} 
\label{tab:hilora_ablation}
\renewcommand{\arraystretch}{1.45}
\setlength{\tabcolsep}{6pt}
\resizebox{\columnwidth}{!}{
\begin{tabular}{ccc cc cc}
\toprule
\multirow{2}{*}{\centering\shortstack{\textbf{Hierarchical}\\\textbf{LoRA}}} &
\multirow{2}{*}{\centering\shortstack{\textbf{LoRA subspace}\\\textbf{clustering}}} &
\multirow{2}{*}{\centering\shortstack{\textbf{Orthogonality}\\\textbf{loss}}} &
\multicolumn{2}{c}{\textbf{CIFAR-100}} &
\multicolumn{2}{c}{\textbf{DomainNet}} \\
\cmidrule(lr){4-5}\cmidrule(lr){6-7}
& & & \textbf{Per.} $\uparrow$ & \textbf{Gen.} $\uparrow$ & \textbf{Per.} $\uparrow$ & \textbf{Gen.} $\uparrow$ \\
\midrule
\checkmark &            &            & 89.7 & 87.5 & 85.4 & 84.2 \\
\checkmark & \checkmark &            & 92.8 & 92.2 & 86.0 & 85.1 \\
\rowcolor{gray!5} 
\checkmark & \checkmark & \checkmark & \textbf{94.1} & \textbf{94.0} & \textbf{87.7} & \textbf{86.1} \\
\bottomrule
\end{tabular}}
\vspace{-4.8mm} 
\end{table}
\noindent \textbf{Ablation Study of HiLoRA Components.}
As shown in Table \ref{tab:hilora_ablation}, by sequentially adding LoRA subspace clustering and cross-layer orthogonality constraints, the performance consistently improves on both CIFAR-100 and DomainNet. In vanilla clustering, we form client groups by running k-means on full parameter updates using cosine similarity \cite{duan2021fedgroup}. We observe that subspace clustering improves client alignment within clusters, and the orthogonality constraints further mitigate interference across tiers.

\noindent\textbf{Cluster Structure and Orthogonality.}
On CIFAR-100, we compute t-SNE embeddings from Jensen–Shannon distances of client label distributions, as shown in Figure~\ref{fig:gsac-tsne}, where clients form compact and meaningful subgroups. On DomainNet, Figure~\ref{fig:hilora-orth-singlecol}\subref{fig:pa-root-leaf} and \subref{fig:pa-cluster-leaf} show the principal-angle distributions between the Root–Leaf and Cluster–Leaf LoRA subspaces, indicating that orthogonality regularization reduces overlap and improves performance.

\noindent\textbf{Generalization Across Adaptation Epochs.} According to the LoRA-subspace similarity, each new client is assigned to its nearest cluster. Loading only the corresponding Root and Cluster adapters achieves 0.811 on CIFAR-100 and 0.849 on DomainNet, surpassing the second-best baseline, which requires 5 epochs of tuning, with 0.788 and 0.840. With 5 epochs of leaf-level updates, the accuracy further improves to 0.859 and 0.861, while cross-client variance continues to decrease. As shown in Fig.~\ref{fig:adapt_epochs}, these results demonstrate that HiLoRA effectively aligns new clients with suitable clusters and reuses their learned parameters, enabling strong generalization with few adaptation steps.

\vspace{-1.2mm} 
\section{Conclusion}

We propose HiLoRA, a multi-level LoRA personalized FL method. We demonstrate its effectiveness through both experimental and theoretical studies. Future work will explore assigning different ranks, as well as extending HiLoRA to LoRA-MoE architectures.

\section*{Acknowledgements}

The above work was supported in part by the Joint Funds of the National Natural Science Foundation of China under Grant U25A20436, Guangxi Key Research \& Development Program (FN2504240036,2025FN96441087), the National Natural Science Foundation of China (NSFC) (62372047,62302049), Guangdong S\&T Programme (No. 2025B0101120006), the Natural Science Foundation of Guangdong Province (2024A1515011323), the Supplemental Funds for Major Scientific Research Projects of Beijing Normal University, Zhuhai (ZHPT2023002), the Fundamental Research Funds for the Central Universities, Guangdong Province Educational Science Planning Research Project under 2025JKZG069, and Higher Education Research Topics of Guangdong Association of Higher Education in the 14th Five-Year Plan under 24GYB207. We also acknowledge the support of the Interdisciplinary Intelligence Super Computing Center of Beijing Normal University at Zhuhai.
\end{CJK}

{
    \small
    \bibliographystyle{ieeenat_fullname}
    \bibliography{main}
}

\end{document}